\documentclass[letterpaper, 10 pt, conference]{ieeeconf}

\IEEEoverridecommandlockouts
\usepackage[pdftex]{graphicx}
\usepackage{balance}
\usepackage{xfrac}
\usepackage{amsmath}
\usepackage{xcolor}

\usepackage[ruled,vlined]{algorithm2e}
\usepackage{subfig}
\usepackage{courier}
\usepackage[per-mode = symbol]{siunitx}

\newcommand{\ms}[1]{\SI{#1}{\meter\per\second}}
\newcommand{\red}[1]{#1}

\begin{document}

\title{\LARGE \bf
Neural-Geometric Tunnel Traversal:\\ Localization-free UAV Flight with Tilted LiDARs
}

\author{Lorenzo Cano, Alejandro R. Mosteo and Danilo Tardioli% <-this % stops a space
\thanks{D. Tardioli is with Centro Universitario de la Defensa, Zaragoza and Instituto de Investigación en Ingeniería de Aragón. A.R. Mosteo and L. Cano are with the University of Zaragoza and Investigación en Ingeniería de Aragón.
Email: {\tt\footnotesize \{dantard, lcano, amosteo\}@unizar.es}}%
}

\thispagestyle{empty}
\pagestyle{plain}

\maketitle

\begin{abstract}

Navigation of UAVs in challenging environments like tunnels or mines, where it is not possible to use GNSS methods to self-localize, illumination may be uneven or nonexistent, and wall features are likely to be scarce, is a complex task, especially if the navigation has to be done at high speed.
In this paper we propose a novel proof-of-concept navigation technique for UAVs based on the use of LiDAR information through the joint use of geometric and machine-learning algorithms.
The perceived information is processed by a deep neural network to establish the yaw of the UAV with respect to the tunnel's longitudinal axis, in order to adjust the direction of navigation. Additionally, a geometric method is used to compute the safest location inside the tunnel (i.e. the one that maximizes the distance to the closest obstacle). This information proves to be sufficient for simple yet effective navigation in straight and curved tunnels.

\end{abstract}

\IEEEpeerreviewmaketitle

%%%% INTRODUCTION %%%%

\section{Introduction}
Robotics in underground environments is a challenging task due to multiple problems that do not arise in outdoors environments. Often, especially in search and rescue tasks, it is not possible to rely on pre-constructed maps; GNSS systems do not work and oftentimes common localization techniques (i.e., scan‐matching, visual SLAM, etc.) do not function as expected due to the lack of usable features or to the absent or irregular illumination \cite{tardioli2019ground}.
However, depending on the application, a precise localization is not indispensable 
for local navigation, and high-level planning may also work without it if other kinds of localization methods exist, such as intersection detection or radio frequency propagation analysis~\cite{seco2022robot}.

In this paper we propose a navigation technique for UAVs that relies on 
the idea of not attempting a precise localization, as we previously demonstrated in~\cite{tardioli2016robot}.
Instead, we propose that the drone can travel at high speed along a tunnel simply knowing its relative position with respect to the axis of such a tunnel.
In other words, we address the case in which the main task of the UAV is to go from a point in a straight or bent tunnel to another at maximum possible speed, leaving to higher-level algorithms the detection of stop triggers, managing bifurcations, etc.

To achieve this objective, we propose to use the readings coming from two 2D LiDARs, one of them mounted on top of the drone and partially tilted upward (thus looking to the ceiling ahead) and another mounted below, tilted downward (and thus looking at the floor ahead). 

These readings are used to estimate the orientation of the UAV with respect to the tunnel walls through a simple and fast deep-neural-network. With this information it is possible to adjust the orientation of the drone to align it with the longitudinal axis of the tunnel and allow a forward displacement. Additionally, the LiDAR points are used to obtain the cross section of the tunnel (projecting them on a vertical plane) and to compute, through a simple geometric technique, the safest spatial position in the tunnel section, intended as the position in which the drone is surrounded by the largest amount of free space in all directions.

The paper is organized as follows: section \ref{sec:related} presents related works. Section \ref{sec:rationale} introduces the rationale behind the proposal while section \ref{sec:details} explains its details. Section \ref{sec:evaluation} presents the results and finally section \ref{sec:conclusions} discusses results and presents the conclusions.

\section{Related works}
\label{sec:related}

Although autonomous drone navigation works are on the rise, they usually do not tackle the particular conditions addressed in this paper, such as poor illumination and a mostly featureless environment. One prolific field is drone racing~\cite{10.1007/s11370-018-00271-6}; however, the demands in this context are focused on path optimization from gate to gate~\cite{10.1109/IROS51168.2021.9636053}, with tracks being well-lit and gates being tagged or, at least, clearly framed~\cite{10.1007/s10514-021-10011-y}. This favors approaches based on computer vision~\cite{10.3390/drones5020052} where neural networks work in relation to gate locations~\cite{10.1109/LRA.2018.2808368},~\cite{10.1007/978-3-030-33749-0_59}.

Moving away from the particular scenario of human-made racing tracks, but still relying on vision sensors, on-board neural networks have also been proven to be effective in unstructured, albeit well-lit, scenarios for path-following~\cite{10.1109/LRA.2015.2509024} and obstacle avoidance~\cite{10.1126/scirobotics.abg5810}. Also, using an off-board neural network has been demonstrated in structured indoors situations~\cite{10.1109/IROS40897.2019.8968222}.

The shortcomings of vision-based methods in poorly-lit conditions~\cite{10.1109/LRA.2020.2969935} have been shown, prompting the use of feature-rich point clouds for localization~\cite{10.1109/ICRA48506.2021.9561335}. The use of LiDAR in combination with neural networks for localization is also proposed in~\cite{10.1109/LRA.2019.2895264}. However, trying to achieve reliable localization in tunnel- and mine-like environments is a notable challenge in itself~\cite{10.1007/978-3-030-95459-8_57}.

In contrast, our work has in common with some vision systems a perception-steer end-to-end pipeline that precludes the need for precise localization or odometry, a feature not observed in these other LiDAR-based examples. Instead, a combination of relative yaw estimation and wall distance detection allows our proposal to safely navigate along the tunnel axis, using a pre-trained neural network. A notable example of improving drone localization by relying on a particular shape available in the environment is found in~\cite{10.1109/MFI52462.2021.9591172}, where the cylindrical shape of a fuselage is RANSAC-extracted from cloud points.

Drone use in underground scenarios is still in the early stages, with many open challenges to autonomy~\cite{10.3390/drones4030034}. For tunnels in particular, the work in~\cite{10.1017/S0263574721000849} presents control solutions also relying on the idea of following the tunnel axis, but with point clouds as input.
In \cite{Sharif2018} the authors propose a navigation approach based on controlling the heading of the UAV, so it advances along the axis of the tunnel using a CNN that classifies the images of the drone as facing to the center, the left or the right of the tunnel. In \cite{Sharif2019} this idea is further refined into a CNN that detects, in each image, the center of the tunnel, allowing finer heading adjustments.
In~\cite{10.48550/arXiv.2201.03312} the focus is placed on airflow effects in very narrow tunnels.

%%%% RATIONALE %%%%

\begin{figure}[!t]
\centering
\includegraphics[width=0.8\columnwidth]{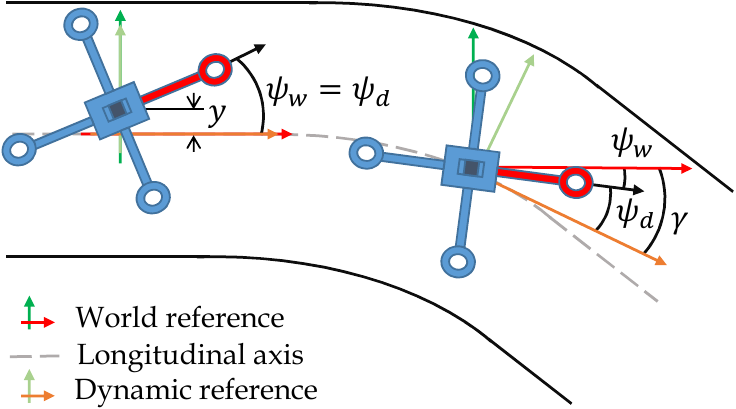}
 \caption{Dynamic reference for UAV navigation. The yaw ($\psi_d$) of the UAV and its $y$ coordinate is expressed with respect to the longitudinal axis. The angles $\gamma$ and $\psi_w$, useful for evaluating the performance, can be obtained from the simulated environment.}
\label{fig:world-dynamic-ref}
\end{figure}

%%%

\section{Rationale}
\label{sec:rationale}
Navigating a tunnel can intuitively seem easy because we, as human beings, are able to obtain the necessary information to walk or drive following, even approximately, its longitudinal axis. Even in the presence of bends, or curves, we are able to adapt and follow the path only with local information adjusting the trajectory in the presence of obstacles. Instead, robots usually  navigate relying on algorithms that need a---precise---absolute localization to move from a point to another. 

For example, if we consider a straight tunnel one-hundred meters long and we need a ground robot to go from the beginning to the end of the tunnel,  we need to define a reference frame first and then provide a pose command relative to that very frame.
Let us assume we place the reference with its origin at the beginning of the tunnel with the $x-$axis coincident with the longitudinal axis of the tunnel and send a command to go to $(x,y)=(100,0)$ to the robot. 
Assuming that the localization system works correctly, the robot will eventually reach the destination.
However, if the tunnel is not completely straight, and instead it is a quarter of a circle of the same length, then the destination coordinates should be adjusted as $(x,y)=(\frac{200}{\pi},\frac{200}{\pi})$. Thus, even though the task is basically the same (reaching the end of the tunnel) the problem is, from a classical point of view, very different. In the first case the robots just need to move forward maintaining its orientation, in the second case it needs to continuously adjust its orientation to match the shape of the tunnel.

\begin{figure}[!t]
\centering
\includegraphics[width=0.8\columnwidth]{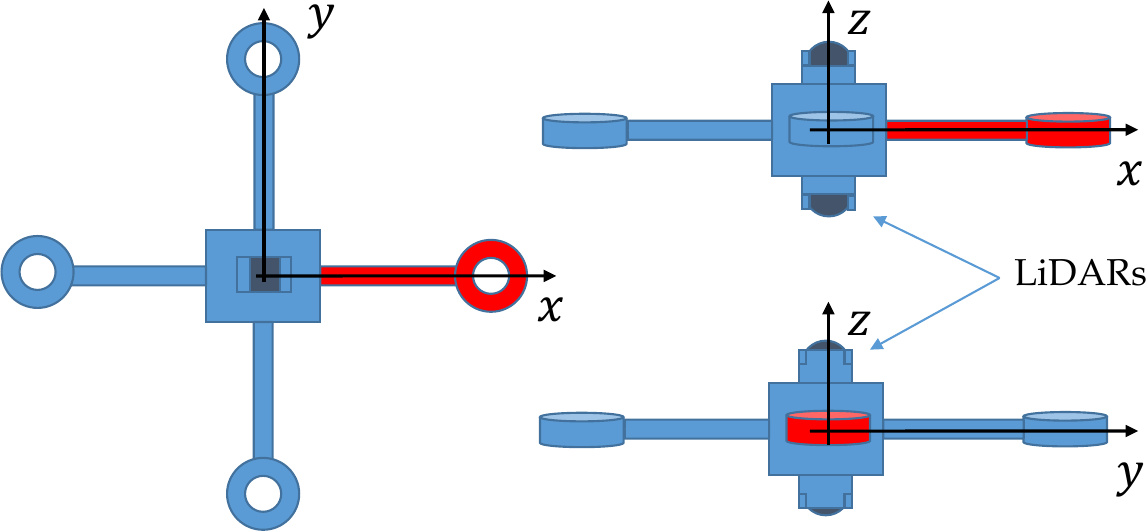}
 \caption{UAV with LiDARs on top and below.}
\label{fig:xyz-view}
\end{figure}

We propose here a different approach, relying on an idea we introduced in~\cite{tardioli2016robot}, in which the position of a robot is given with respect to the tunnel itself: the $x$ coordinate is given by the distance traveled by the robot from the beginning of the tunnel, the $y$ coordinate corresponds to the displacement from the longitudinal axis of the tunnel, and the orientation is given by the angle between the robot's own frame and the tunnel longitudinal axis tangent. Consider Fig.~\ref{fig:world-dynamic-ref}: the actual UAV yaw is $\psi_w$; however, according to the dynamic reference definition just given, it would be $\psi_d$.

This way, the two tasks described above become essentially the same one, which is going to coordinates $(100,0)$, and that can be solved in a way which is closer to how a human would.

Specifically, the robot should continuously adjust its orientation to that of the tunnel longitudinal-axis tangent and move forward until it reaches the end of the tunnel. Ultimately, everything boils down to the estimation of the UAV's orientation with respect to the longitudinal axis of the tunnel.

If the tunnel does not have very sharp curves or irregular walls, it could be possible to estimate the orientation by taking advantage of the wall LiDAR readings (for example, using the Hough transform, as done in~\cite{tardioli2016robot}). Also, even though ground robots can have a consistent reading of the horizontal plane, UAVs have a harder time due to their larger number of degrees of mobility. Namely, rotations in roll ($\phi$) or pitch ($\theta$) can cause the drone to temporarily lose view of the walls.

Thus, we propose an alternative solution using a light neural network based on the same idea of detecting the orientation from the LiDAR wall readings, but without those drawbacks caused by UAV specifics or irregular surroundings.

\begin{table}[ht]
  \centering
  \begin{tabular}{|l|c|c|c|c|}
    \hline
    Model           & Type & Price (USD)    & Weigth  & Power\\
    \hline
    UST-10LX        & 2D   & $\approx$ 1750 & 130g    & 1.3 W  \\
    \hline
    Puck Light      & 3D   & $\approx$ 5500  & 590g    & 8 W\\
    \hline
  \end{tabular}
  \caption{Comparison between Hokuyo UST-10LX 2D and Velodyne Puck Light 3D LiDARs.}
  \label{tab:lidar-comparison}
\end{table}

\begin{figure}[!t]
\centering

    \centering
    \hfill
    \subfloat[]{{\includegraphics[height=1.5cm]{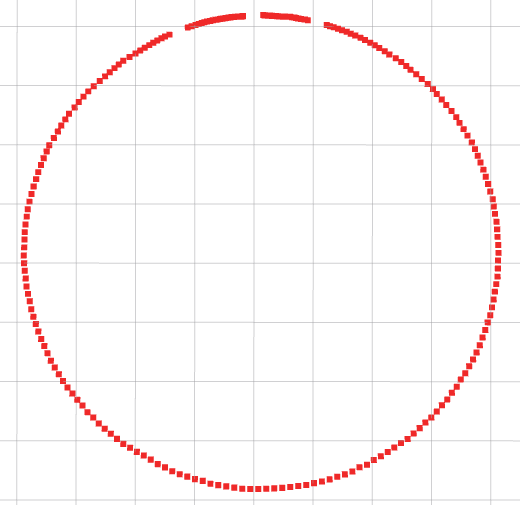} }}%
    \hfill
    \subfloat[]{{\includegraphics[height=1.5cm]{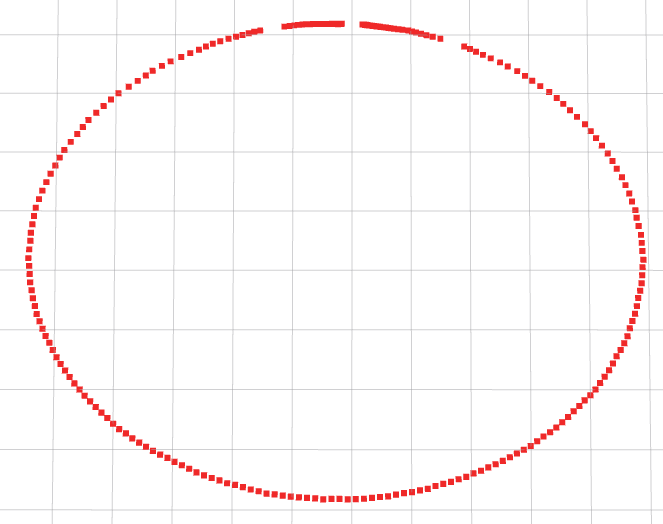} }}%
    \hfill
    \subfloat[]{{\includegraphics[height=1.5cm]{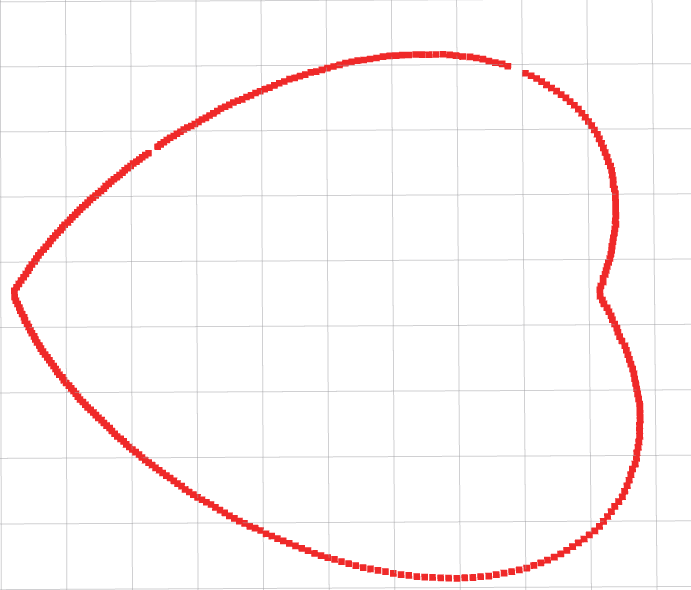} }}%
    \hfill
    \subfloat[]{{\includegraphics[height=1.5cm]{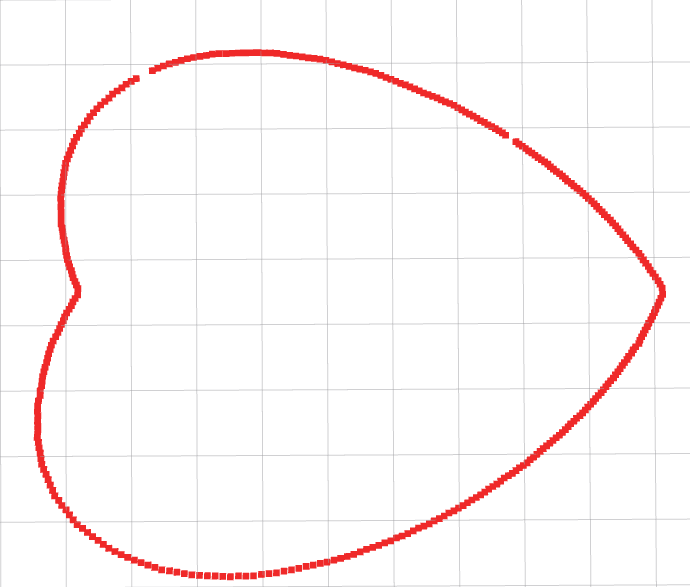} }}%
 \hfill
 \caption{LiDAR readings deformation in a cylindrical pipe. Vertical LiDARs with null yaw  (a). Vertical LiDARs with not null yaw (clockwise and counterclockwise) (b). Tilted LiDARs with non-null yaw (clockwise) (c). Tilted LiDARs with non-null yaw (counterclockwise) (d).}
\label{fig:deformation}
\end{figure}

%% tilted lidars %%

\begin{figure}[!b]
\centering
\includegraphics[width=0.55\columnwidth]{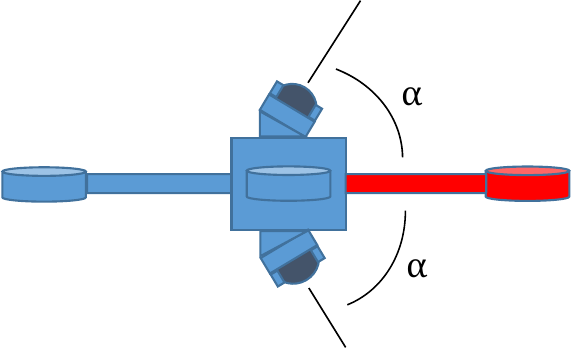}
 \caption{UAV with tilted LiDARs.}
\label{fig:inclined}
\end{figure}

\subsection{Tilted LiDARS}
\label{sec:tilted}

Consider an UAV equipped with two LiDARs, one of them mounted on top and pointing squarely upward and the other mounted symmetrically below and pointing downward (see Fig.~\ref{fig:xyz-view}). If this UAV is placed inside of a pipe at mid height and with its frame $x$-axis aligned with the pipe longitudinal axis, then the LiDAR readings, both lying on the vertical $y-z$ plane, will draw together the exact front shape of the pipe, as cut by that plane (i.e. a circle, see Fig.~\ref{fig:deformation}.a). Now, if we slightly change the yaw orientation of the drone, the LiDARs will draw a different shape, an ellipse, which is the result of an oblique cut (see figure \ref{fig:deformation}.b).

Now, given such an ellipse, it is possible to compute the orientation that generated it instead of the circle we would expect when perfectly aligned with the pipe axis, and thus obtaining, ultimately, the orientation of the drone. 
This can be done, for example, by rotating a plane around the drone's $z-$axis and projecting the points on it until a circle appears.

However, this idea has an important drawback: a yaw rotation of any pair of angles $\alpha$ and $-\alpha$ will result in the same ellipse, thus giving two possible solutions for every $\alpha\neq 0$.

To solve this problem, it is possible to install the LiDARs in a way that the LiDAR-readings plane has a known inclination with respect to the drone's $x-y$ plane (see Fig.~\ref{fig:inclined}). In this way, if the readings are projected to the same $y-z$ plane, they will draw a circular shape when the UAV has a null yaw, but will deform to now different figures when the UAV rotates clockwise and counterclockwise as shown in figures \ref{fig:deformation}.c) and d). The analysis of these figures can yield the yaw of the UAV in the same way as discussed above but this time without the uncertainty about the sign of the yaw angle.

\subsection{Retrieving the Yaw}
\label{sec:retrieving}

To obtain the yaw angle we can intuitively rotate a vertical plane around the $z-$axis until a circle appears. However, computationally speaking, detecting that the circle has actually appeared is not an easy task, especially if we consider that the tunnel will  usually not have a perfectly circular section, rather having a horseshoe-like shape. To solve this problem, we used a machine-learning based approach that will be explained in Sec.~\ref{sec:details}. 

\subsection{Roll and Pitch}
Until now, we proposed to use the projection of the LiDARs readings on the UAV's $y-z$ plane to estimate the yaw rotation implicitly assuming that this plane is vertical (i.e., parallel to the force of gravity). What happens, though, when the UAV also rotates around the roll and pitch axes?
Unlike yaw, roll and pitch can be precisely estimated using inclinometers, often integrated in IMU units. \red{Inclinometers exclusively depend on gravity to compute orientations, ensuring they avoid error accumulation seen in IMUs. As a result, they are reliable for estimating a plane aligned with gravity's force which allows direct compensation for deformation caused by rotation around these two axes.} For example, if the UAV rotates around the pitch axis by an angle $\alpha$, we can easily rotate the LiDAR readings by $-\alpha$ before projecting them on the vertical plane. We can proceed likewise for the roll.

\subsection{Safe navigation}
\label{sec:ratsafe}

Estimating the UAV's yaw is only part of the task. It is also necessary to move along the tunnel without crashing---at least---against the wall. 

The proposed navigation scheme consists in adjusting the orientation of the UAV accordingly to the direction of the longitudinal axis of the tunnel  while moving forward safely, as explained above.

We assume that the safest way is to maintain the UAV as far as possible from the tunnel walls (and from hypothetical objects) at all times while moving in the cited direction.
However, this means that it is necessary to estimate the position of the UAV with respect to the tunnel longitudinal axis.
To do so, we process again the LiDAR readings. After obtaining the UAV's yaw, we obtain the tunnel section by rotating the LiDAR point cloud around the yaw, roll and pitch axes of the corresponding angles (notice that roll and pitch can be obtained directly from the inclinometer as explained earlier). Second, we use a fast algorithm (explained in Section~\ref{sub:circles}) to compute the largest circle that fits within such a section. We assume that the center of the circle is the safest position for the UAV to be at. Given that the circle is obtained in terms of the point-cloud, which in turn is represented on the UAV's reference frame, the center of the circle is expressed in the UAV-frame coordinates. This means that we can easily calculate the vector that indicates the direction the UAV needs to move toward in order to be in the safest position; \red{applying the correspondent $y$ and $z$ speed the drone will move toward the safest position computed. }

Iteratively adjusting the UAV's yaw angle and position into the tunnel section while forcing a forward velocity, allows the drone to navigate along. \red{Such forward velocity is determined by a maximum value, typically contingent on the intricacy and spatial breadth of the surroundings and on the width of the circle computed using as described just above: when the circle is smaller than a certain threshold the speed is reduced accordingly up to a minimum speed.}

In the next sections a detailed explanation of the process and the specific techniques used will be presented together with simulation results and considerations.

\section{Proposal Details}
\label{sec:details}

\begin{figure}[!t]
\centering
\includegraphics[width=0.95\columnwidth]{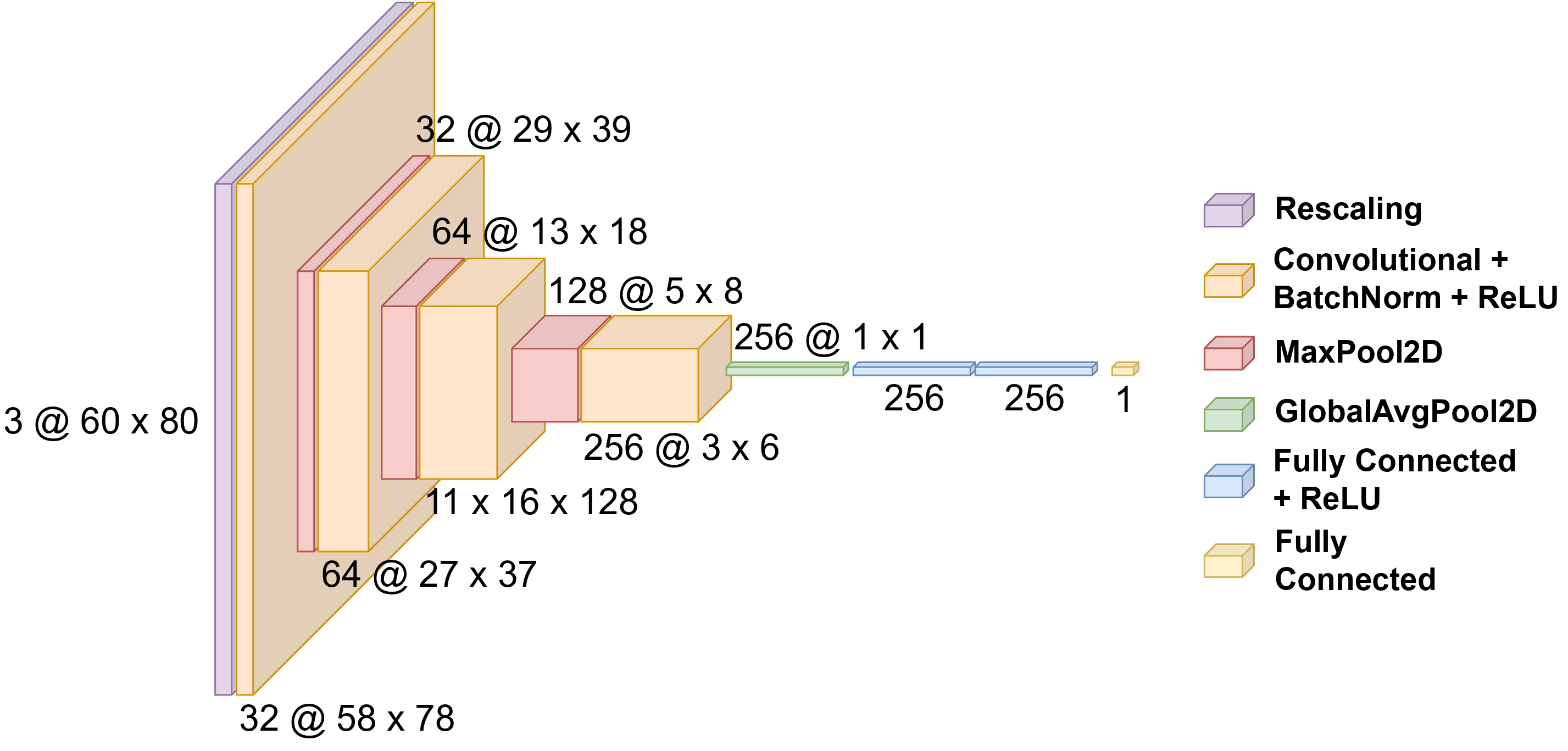}
\caption{Diagram of the CNN architecture.}
\label{fig:network-diagram}
\end{figure}

As discussed in Sec.~\ref{sec:tilted}, the readings coming from the tilted LiDARs have a different shape if the UAV is rotated clockwise and counterclockwise, and the deformation is more extreme when the absolute value of the yaw is larger. This information seems enough for a well trained network to be able to infer the UAV's orientation.

\red{
The proposed CNN is implemented using the PyTorch library following the scheme described in Fig.~\ref{fig:network-diagram}. We used Mean Squared Error as loss function and an RMSProp optimizer. 
}

\begin{figure}[b!]%
    \centering
    \subfloat[]{{\includegraphics[width=0.495\columnwidth]{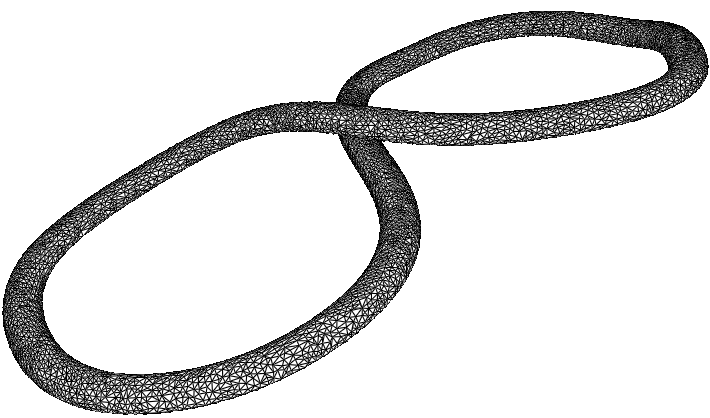} }}%
    \subfloat[]{{\includegraphics[width=0.36\columnwidth]{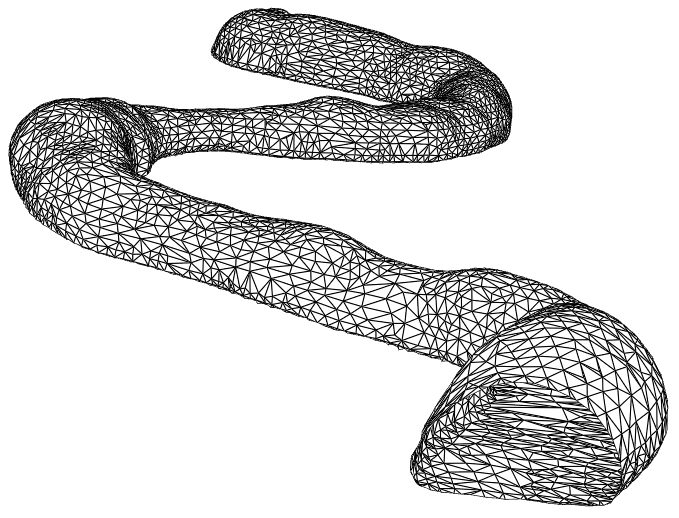} }}%
 \caption{Examples of tunnel meshes created with the generation tool. Eight-shaped 3D tunnel with no roughness and a radius of \SI{2}{\meter} (a) and S-shaped tunnel with roughness $0.1$ and average radius of  \SI{4}{\meter} (b).}
\label{fig:environment}

\end{figure}

\begin{figure}%
    \centering
    \subfloat[]{{\includegraphics[width=1.6cm]{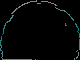} }}%
    \subfloat[]{{\includegraphics[width=1.6cm]{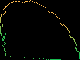} }}%
    \subfloat[]{{\includegraphics[width=1.6cm]{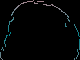} }}%
    \subfloat[]{{\includegraphics[width=1.6cm]{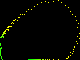} }}%
    \subfloat[]{{\includegraphics[width=1.6cm]{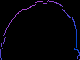} }}%
    \caption{Example of CNN training images}%
    \label{fig:net-input}%
\end{figure}

We fed the network with images generated in the following way. First, the LiDAR readings are converted to a point cloud represented on the stabilized frame mentioned earlier and each point is projected to the drone's vertical plane obtaining a 2D $y-z$ representation on the drone's frame. The image is then scaled and cropped to fit within a 60x80 pixel surface. Each pixel is successively painted with a color derived from the $x-$coordinate (i.e., the distance from the origin) of the original point, the pitch, and the roll of the UAV; the latter two obtained through reading the on-board inclinometers. Some examples of images used as network input are shown in Fig.~\ref{fig:net-input}.

The output of the network would eventually be the value of the yaw estimated for the UAV. This information is then used in two ways. On the one hand, a proportional controller is used to generate a rotation velocity on the yaw axis to force the UAV to reestablish the correct orientation (i.e. aligning the UAV's $x-$axis with the tunnel longitudinal axis). On the other hand, the estimation is also used to rotate the point cloud by the same amount to obtain the \textit{real} cross section of the tunnel and computing the safest spots and, in turn, the corresponding $y$ and $z$ velocities.

\subsection{Training of the Network}
The network was trained with LiDAR readings obtained through simulation in hundreds of real-world-like tunnels.

\subsubsection{Tunnel generation}
\label{sec:procedural}
The tunnels were generated using the tool we developed and made available in \cite{cano2023procedural}. This tool allows the creation of arbitrarily complex tunnels and tunnel networks with tuneable diameter and level of roughness that can be easily imported in Gazebo or any simulation tool capable of importing meshes. The tool has the option of generating tunnels through user-defined graphs, or randomly by specifying a set of generation parameters.
Figure \ref{fig:environment} shows two examples of tunnels generated using the cited tool with different levels of roughness.

\subsubsection{UAV Setup}
The LiDARs were mounted with an inclination of $\pm\SI{45}{\degree}$, an angular range of $\pm$\SI{180}{\degree}, a resolution of~\SI{1}{\degree} and an update rate of \SI{60}{\hertz}. 

The tilt angle was chosen to be a middle ground: with too small an angle, readings might be too far away to be representative of the close surroundings of the drone; conversely, a large angle produces distortions in the shapes that are negligible, especially when the yaw is small.

\subsection*{Training process}
For the training, a batch process was set up in which 1) a random tunnel is generated with a certain level of roughness, diameter and shape, 2) a random $\psi_d$ (refer to Fig.~\ref{fig:world-dynamic-ref}) is chosen in the range of $[-40,40]$ degrees, 3) the UAV is artificially placed (using Gazebo dedicated topic) in a random position inside such tunnel setting random $y$, $z$ (such that the drone is never more than 2 meters away from the axis of the tunnel), roll and pitch (both in the range of $[-20, 20]$ degrees) values and the yaw correspondent to the $\psi_d$ previously chosen (notice that knowing $\gamma$ from the simulated environment it is trivial to compute $\psi_w$ from $\psi_d$), 4) a LiDAR reading is captured, the image created and fed to the CNN using $\psi_d$ as label for that image.
Steps 2 to 4 are repeated several hundreds times for the same tunnel before the algorithm loops back to step 1. Notice that the tunnel generation tool provides, for each tunnel generated, the position and the tangent of each point of its longitudinal axis which makes it trivial to compute the absolute yaw from the $\psi_d$ and vice versa.

In this way, we collected about 3.1GB of data or about 120k samples. We used $90\%$ of the data for training and $10\%$ for validation.

\begin{figure}[!t]
\centering
\includegraphics[width=0.95\columnwidth]{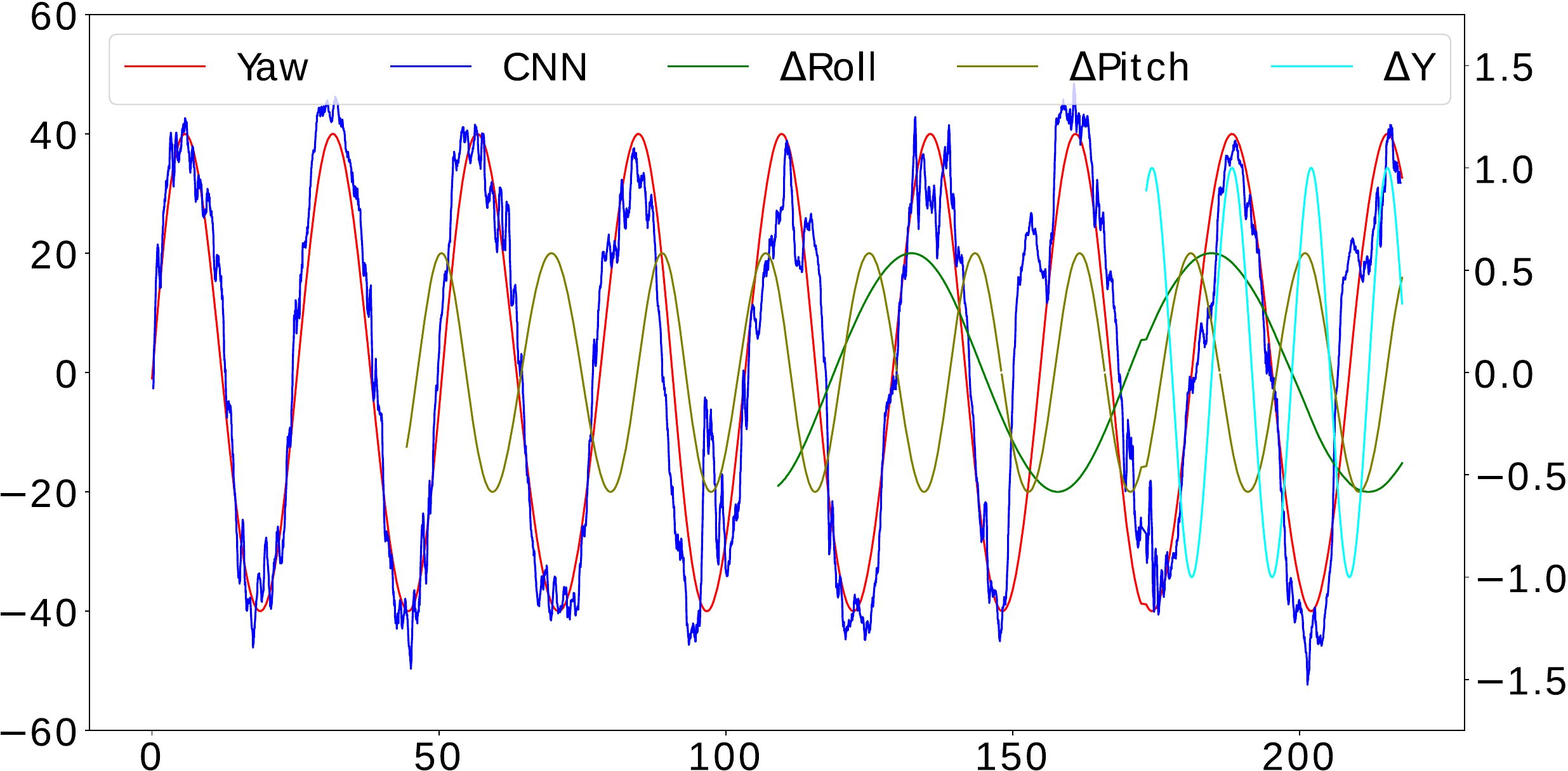}
 \caption{CNN output obtained moving the drone artificially in a simulated environment. Yaw ground truth with respect to the tunnel longitudinal axis (red), roll and pitch ground truth (greens), $y$ position (cyan) and network output (blue).}
\label{fig:net-results}
\end{figure}

\subsection{Estimating the Safest Spot and Speed}
\label{sub:circles}

The second leg of the proposed navigation algorithm is the continuous estimation of the safest spot within the tunnel. As advanced in Sec.~\ref{sec:ratsafe}, we assume that after every LiDAR scan and yaw estimation, we obtain a projected 2D point cloud that represents the section of the tunnel. We want to estimate the point within the section whose minimum distance from the closest point of the section is maximum or, in other words, the largest circle than can be inscribed within the polygon represented by the tunnel section, whose center can be considered the safest location for the drone.
To do so, we use an algorithm named \textit{polylabel}, based on a method described in \cite{poleofinaccessibility}, improved by Vladimir Agafonkin, whose code is available at \cite{polylabel}.
%https://www.tandfonline.com/doi/abs/10.1080/14702540801897809

This method uses quadtrees to recursively subdivide a two-dimensional space into four quadrants and recursively subdivide them into four smaller cells, probing cell centers as candidates and discarding cells that cannot possibly contain a solution better than the one already found.
Since the search is exhaustive, the algorithm will eventually find a cell that is guaranteed to be within a global optimum. 

We apply this algorithm to the adjusted section of the tunnel obtaining the pole of inaccessibility, that is, the center $c=(y_c,z_c)$ of the largest inscribed circle $C$ and its radius $r_c$. We would like that point to be the position of the drone at any given moment. This point is found on the UAV's reference frame (that is, the drone is always in position \mbox{$(x,y,z)=(0,0,0)$} in that frame); thus, we can easily calculate the velocities $v_y=k_y \cdot y_c$ and $v_z=k_z\cdot z_c$ to be applied to the drone to move it to $(y_z, z_c)$, being $k_y$ and $k_z$ two proportional controller constants. To speed up calculations, we reduce the complexity of the input polygon (that would otherwise have as many edges as the number of points of the point cloud) using the Ramer-Douglas-Peucker algorithm \cite{douglas1973algorithms} before providing it to the \textit{polylabel} algorithm. In this way, the number of edges is reduced by, at least, one order of magnitude.

\begin{figure}[!t]
\centering
\includegraphics[width=0.67\columnwidth]{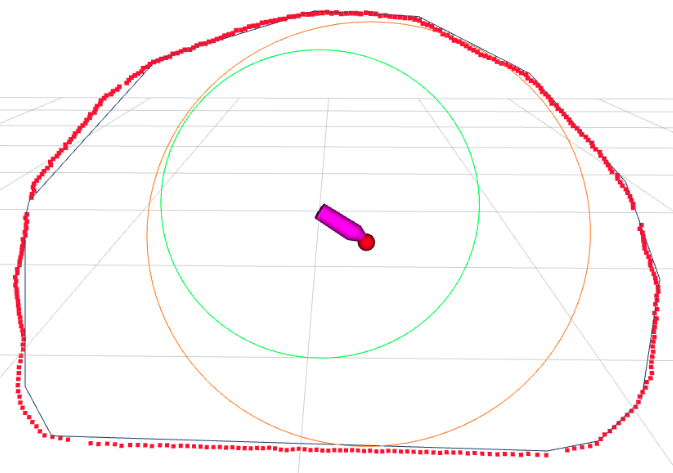}
\caption{Adjusted Point Cloud (red), Ramer-Douglas-Peucker (blue), maximum inscribed circle (orange) and its center (red), safety circle (green), $y-z$ velocities (magenta vector).}
\label{fig:circles}
\end{figure}

\red{
To make the navigation safer in situations in which the environment has an irregular section (for example a cave), we allow a variable $x$-axis speed $v_x$ dependent on the environment itself.
Specifically, the maximum speed $v_{max}$ is scaled according to the ratio between the radius $r_c$ of the circle $C$ computed at each LiDAR reading event and the radius $r_{sc} < r_c$ of the \textit{safety circle} $C_{safety}$ defined as a circle around the UAV on its $y-z$ plane (see Fig.~\ref{fig:circles}) whose fixed radius depends on the user perception of the environment complexity. Also, if this ratio is below a certain threshold, the speed is reduced to a minimum value $v_{min}$. To summarize:
\begin{equation}
v_x = \left\{ \begin{array}{lcc}
             v_{max} &   if  &  R_c \ge 1 \\
             v_{max}\cdot R_c^4 &   if  & th < R_c < 1 \\
             v_{min} &   if  & R_c \le th\\
             \end{array}
   \label{eq:v_max}
   \right.
\end{equation}
being $R_c=\dfrac{r_{sc}}{r_c}$. 
}
Notice that in the second part of the function $v_x$ the relationship is elevated to the fourth power to make the reduction of the speed faster when the available space starts to decrease.

\subsection{Avoiding restless motion}
Given that a LiDAR readings are noisy, the point $c$ of $C$ will be moving slightly, inducing a restless motion of the drone even in static conditions. To avoid that, we established a control algorithm in which as long as the safety circle is within the circle $C$, no changes will be effected to the drone $y$ or $z$ position. As soon as the condition does not hold, the drone is moved applying the corresponding control of velocities $v_y$ an $v_z$.

\section{Evaluation}
\label{sec:evaluation}

\subsection{Implementation}

\begin{figure}[!t]
\centering
\includegraphics[width=0.99\columnwidth]{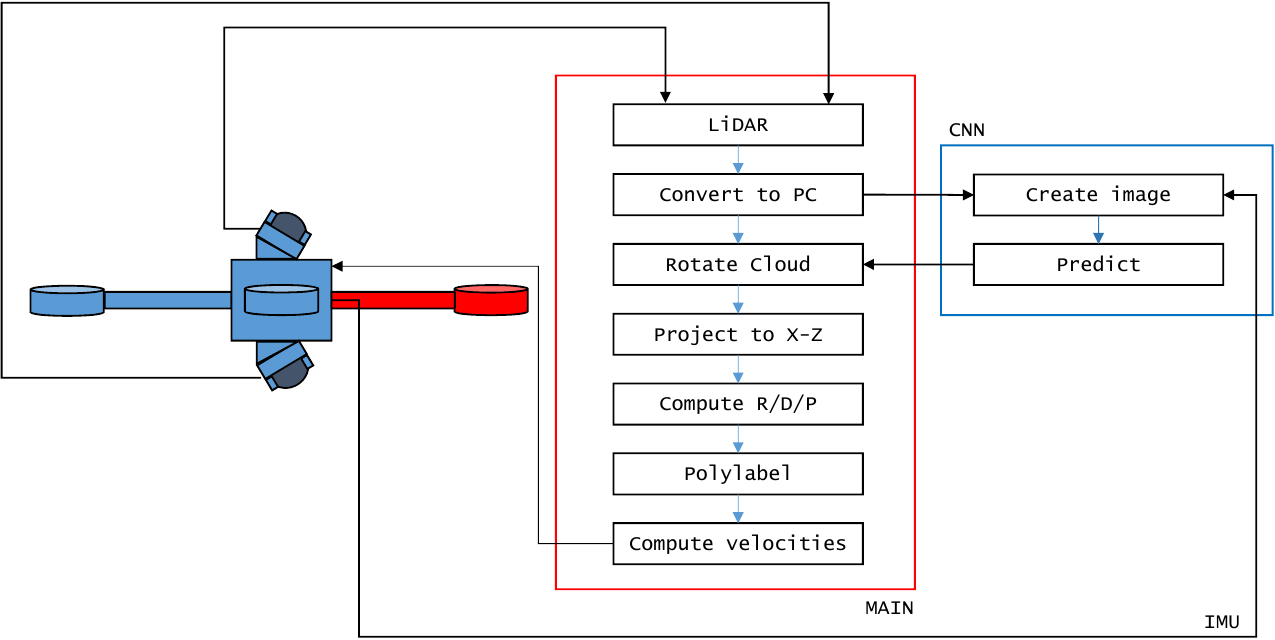}
\caption{ROS nodes used to implement the system and their connections.}
\label{fig:nodes}
\end{figure}

The described solution has been implemented on the ROS platform. The system is composed by two ROS nodes, the first of them (written in C++) in charge of managing sensors, the computation of the safest spot and in general of UAV navigation, with the second node in charge of predicting the yaw using the cited CNN given the point cloud sensed by the LiDARs (this one, written in Python, uses the PyTorch library).
Figure \ref{fig:nodes} shows the detailed scheme of the platform and the connections between both nodes. 

The node $\textrm{MAIN}$ receives readings coming from the LiDARs, converts them to a point cloud and sends it to the CNN node that predicts the UAV's yaw using the cited neural network, using also the pitch and roll readings. The yaw estimation is sent back to the MAIN node, which using this information first rotates the point cloud by the same angle and projects it to the $y-z$ plane, then uses the Ramer-Douglas-Peucker algorithm to reduce the cloud to a simpler polygon and  computes the pole of inaccessibility. Successively, it computes the $v_y$, $v_z$, $v_{yaw}$ velocities, that are applied immediately.

\subsection{Neural Network Performance}
One of the most important aspects of the proposed method is the performance of the CNN used to estimated the UAV yaw. After training it for $400$ epochs and obtaining a satisfactory theoretical performance, we evaluated it in a simulated scenario in which the UAV was moved artificially inside a bent tunnel (never used for training) varying the yaw, roll pitch and $y$ position (intended as in Fig.~\ref{fig:world-dynamic-ref}) in a know way following a sinusoidal function.
Figure \ref{fig:net-results} shows a comparison between the ground truth represented as the real angle of the drone with respect to the axis of the tunnel (i.e. $psi_d$, in red) and the correspondent angle inferred by the CNN (blue) filtered with a moving average to improve visualization (in a real-world situation most of the noise would be filtered out by the UAV's own inertia, too). Green lines show the UAV roll and pitch for each sample, respectively while the cyan line shows the $y$ position. While the yaw is varied from the beginning, the other three parameters start at different moments in time. As it can be seen, despite being somehow noisy the network performs reasonably well even in case of notable variations of roll, pitch and $y$ parameters.

\begin{figure}
\centering
\includegraphics[width=0.78\columnwidth]{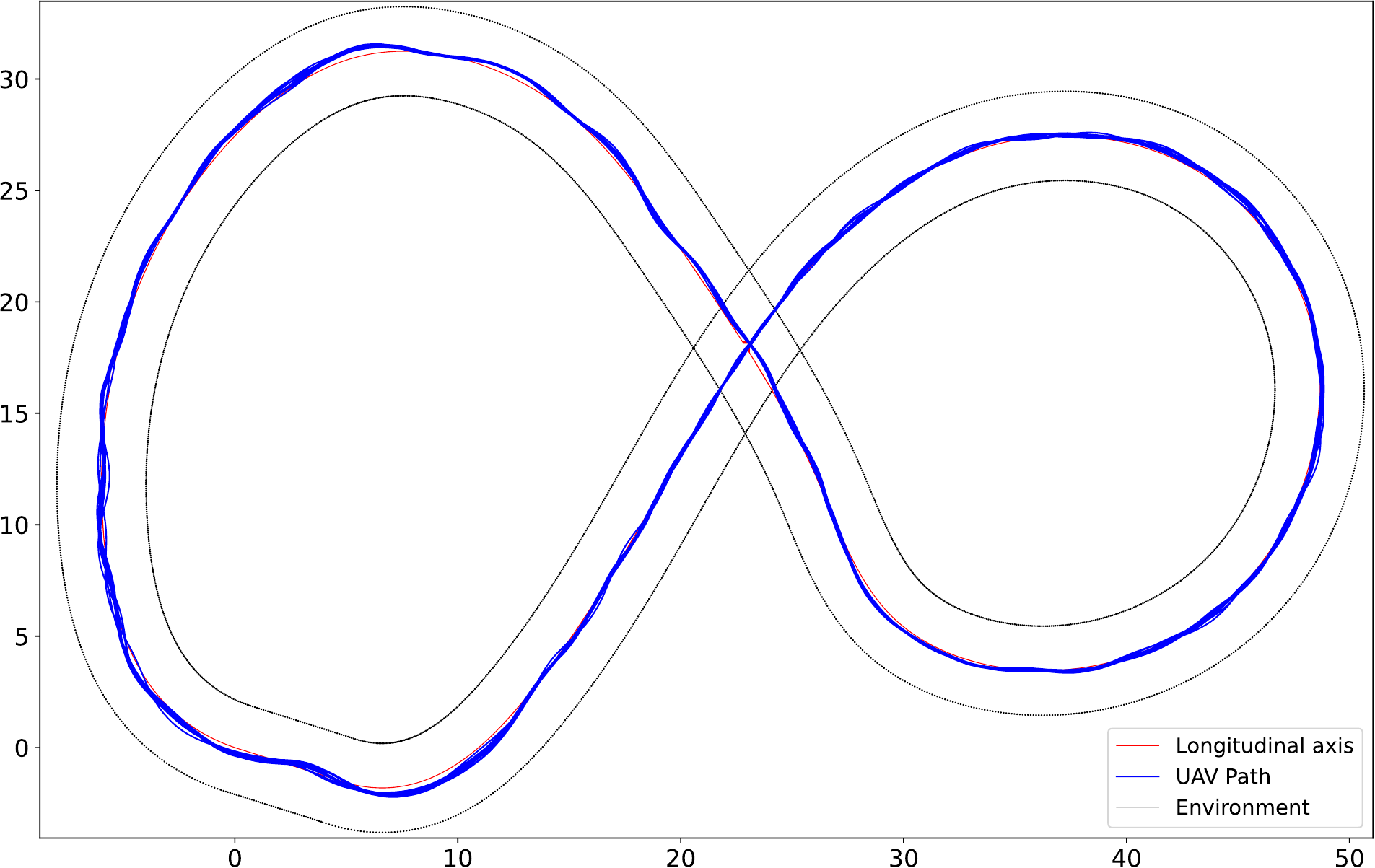}

\caption{Path of the drone in a 8-shaped circuit traveled ten times.}
\label{fig:ocho_10_vueltas}
\end{figure}

\begin{figure*}[]
\centering
\includegraphics[width=1.98\columnwidth]{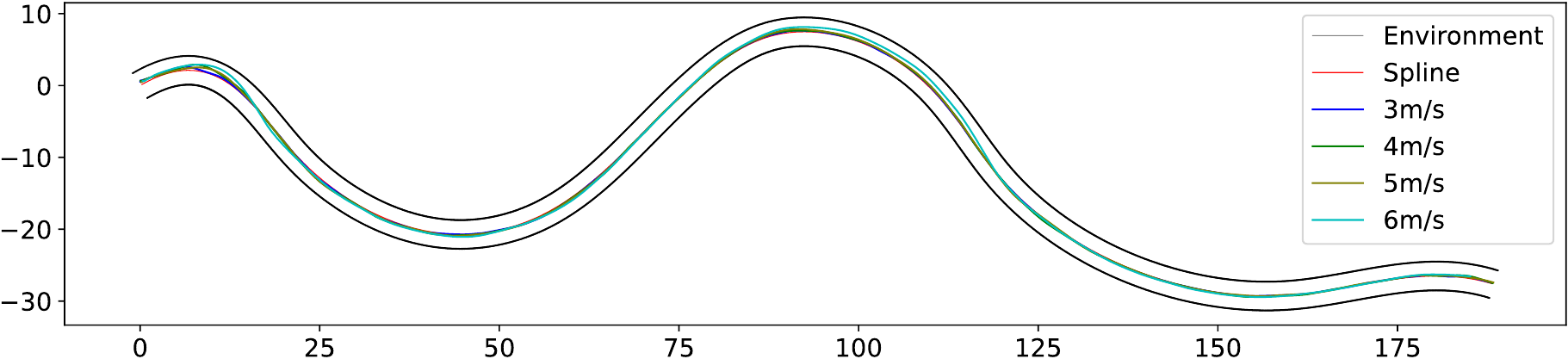}
\caption{Path of the UAV in double-S shaped tunnel with no roughness at different speeds.}
\label{fig:short-plain}
\end{figure*}

\subsection{Simulations}
The testing has been performed using the Gazebo simulator integrated into the ROS platform using the simulated UAV provided by the \texttt{hector\_quadrotor} package \cite{KohlbrecherMeyerStrykKlingaufFlexibleSlamSystem2011} (4 propellers, \SI{80}{\cm} width, inertias (\SI{}{\kilo\gram\per\metre\squared}) $i_{xx}=0.01152$, $i_{xy}=0.0$, $i_{xz}=0.0$, $i_{yy}=0.01152$, $i_{yz}=0.0$ and $i_{zz}=0.0218$) and a set of environments of different shape and roughness generated with the tool briefly described in Sec.~\ref{sec:procedural}. 

\subsubsection{Speed}
\begin{figure}[b!]
\centering
\includegraphics[width=0.95\columnwidth]{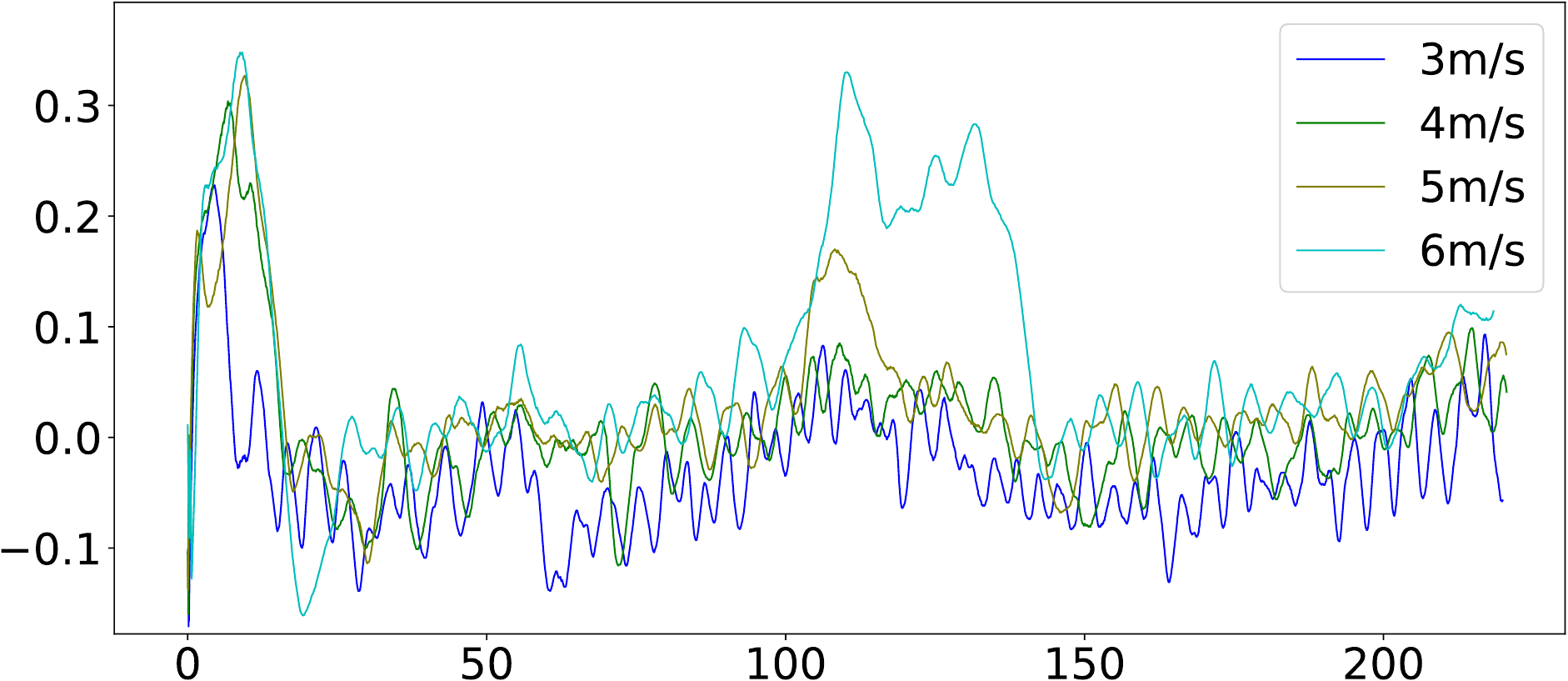}
\caption{Heading error of the UAV at different speeds (radians).}
\label{fig:heading-error}
\end{figure}

Fig.~\ref{fig:short-plain} shows the trajectory of several trials carried out in a double-S-shaped 2-meters-radius smooth tunnel with a horseshoe section at different \textit{constant} speeds between \ms{3} and \ms{6}. As it is possible to see, the drone is capable of traversing the tunnel showing, as expected, a growing error in terms of being able to follow the longitudinal axis, due to the growing centrifugal force that it suffers from in the curves. Fig.~\ref{fig:heading-error} shows the corresponding errors in terms of heading misalignment with respect to the longitudinal axis tangent (i.e., $\psi_d$ error). As shown, and ignoring the beginning of the graph when the drone takes off, the error never exceeds $\pm0.1$ radians (about 6 degrees) until \ms{4} while it grows up to $0.15$ and $0.32$ radians for \ms{4} and \ms{5} respectively. Still, the error is quickly recovered and the orientation is correct for the most part of the trial.

\subsubsection{Consistency}
Fig.~\ref{fig:ocho_10_vueltas} shows the performance of the navigation technique in a 8-shaped circuit shown in the same figure and that the drone has traveled a total of 10 times at a constant speed of \ms{3}. In this case the tunnel had a radius of~\SI{1}{\meter}. Considering that the drone has a width of \SI{80}{\cm} (similar, for example, to a DJI F450 quadcopter), this leaves just a \SI{60}{\cm} gap on each side. As it can be observed, the trajectory is pretty consistent and it shows the reliability of the proposed scheme.

\begin{figure*}[]
\centering
\includegraphics[width=1.96\columnwidth]{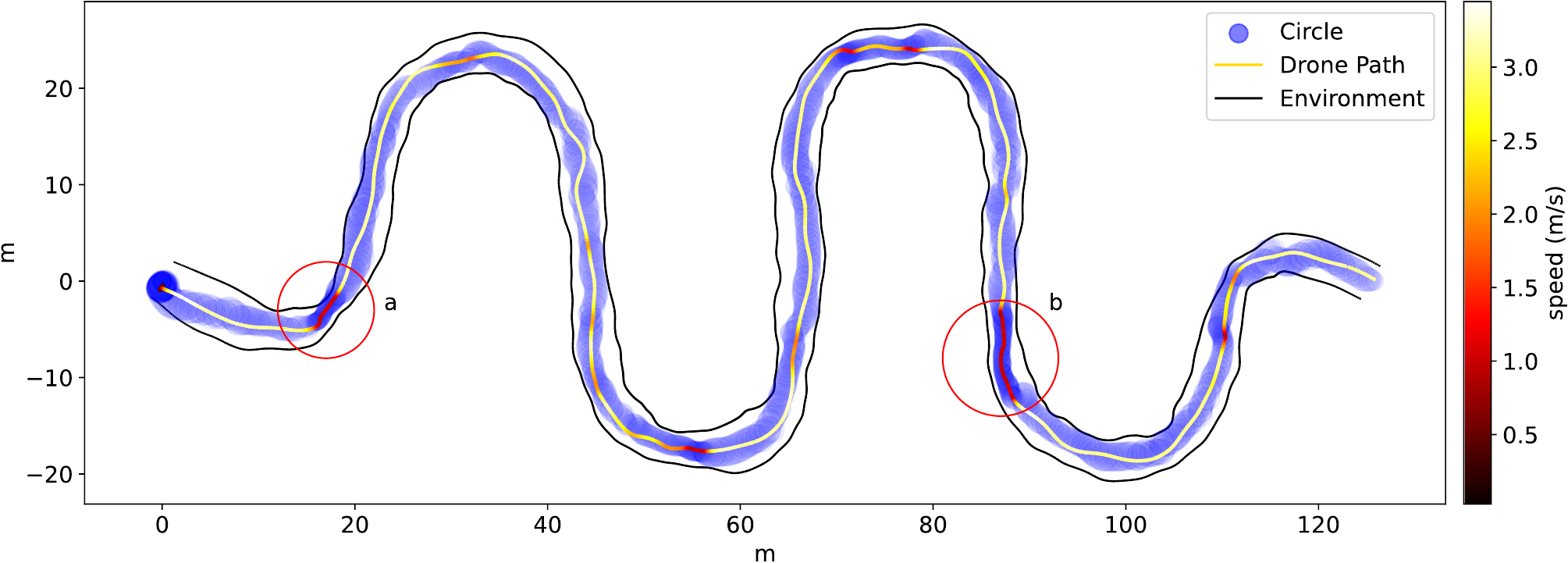}
\caption{Triple-S-shaped rough simulation environment. Circle $C$ computed by the UAV at each point (blue), and its path and speed (color scale). The black line represents by the section of the environment at the height of the tunnel longitudinal axis.}
\label{fig:tough}
\end{figure*}

\subsubsection{Rough environments}
 
We also verified the efficacy in tougher environments, more specifically in tunnels where the surface is rough like in real-world underground mines or caves.

In these environments the shape of the cross section is strongly influenced by irregularities of the internal surface. This has at least two effects. On the one hand, this makes it harder for the network to infer the orientation of the drone. On the other hand, the irregularities may reduce considerably the free space around the drone and, in turn, the diameter of the $C$ circle, making the navigation more challenging. 

\begin{figure}[]
\centering
\includegraphics[width=0.75\columnwidth]{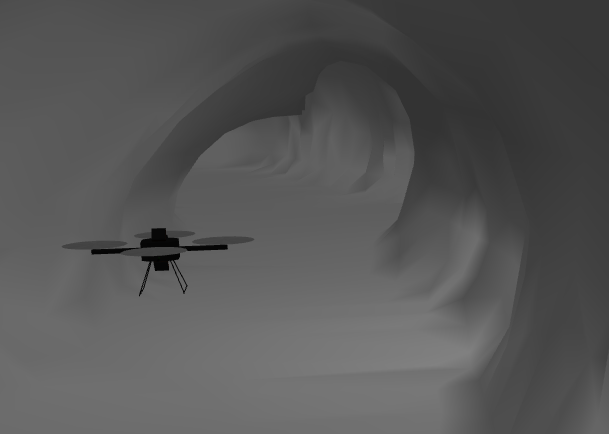}
\caption{Snapshot of a stretch of the Gazebo environment in which it is possible to observe the irregularities that provoke the shrinking of the circle $C$.}
\label{fig:estrecho}
\end{figure}

This is clearly visible in Fig.~\ref{fig:tough} where a triple-S rough environment with sharp turns is shown. Even if it does not seem especially irregular, the reader should consider that the black line represents a $x-y$ section at a specific height while the 3D structure is notably more complex. The figure shows the path of the UAV inside such a tunnel and its speed, that this time is not constant as it will be explained in a moment. The width of the blue brush represents the diameter $d$ of the $C$ circle computed by the UAV at each point while it is traveling. As it can be seen the value of $d$ is not constant and there are zones in which, despite the section of the cited $x-y$ projection appearing to be pretty wide, the space actually available to the UAV for flying is minimal (see for example point b and Fig.~\ref{fig:estrecho}). 

Navigating in this environment requires reducing the speed at certain points to avoid UAV crashes.
Equation \ref{eq:v_max} guarantees that the UAV will travel at a speed compatible with the environment. For example, in Fig.~\ref{fig:tough} it is possible to observe that the speed is adapted to the width of the circle $C$ (which is most of the time larger that the circle $C_{safety}$ allowing the drone to travel at the maximum speed of \ms{3}). However, in points a) and b) the UAV finds itself in a situation in which the speed must be reduced to the minimum, according to the third part of equation \ref{eq:v_max} (\ms{1} in this case).
With this adjustment, the UAV is capable of covering the \SI{260}{\meter} of the environment in approximately \SI{107}{\second} achieving an average speed of \ms{2.4}.

\red{\subsection{Challenges}
Even if the method presented has been trained and tested in a realistic simulator within different procedurally generated tunnels with semicircular and horseshoe shape, where it demonstrated excellent behavior also in strongly degraded tunnels, it would be probably necessary to retrain the network in case of tunnels with strongly different shapes or in highly asymmetric ones. On the other hand, it needs yet to be tested in real-world environments to verify its performance and solve hypothetical problems that could appear in those setting such as motion-induced distortions on the LiDAR measurements. A real-world implementation could also need more advanced controllers than proportional gain controllers. 
}

\section{Conclusion}
\label{sec:conclusions}

This work presented a proof-of-concept method for the navigation
of UAVs in tunnels, relying only on two 2D LiDARs mounted on top and below the UAV body, looking forward and tilted about their pitch axis. These sensors are used for continuous estimation of the UAV's yaw by processing the image obtained from projecting the readings to a vertical plane through the use of a lightweight convolutional neural network. Additionally, a geometric method based on finding the pole of inaccessibility is used to maintain the drone away from the tunnel walls, avoiding crashes. 

Extensive simulations within a bent tunnel show the effectiveness of the proposal for speeds up to \SI{6}{\meter\per\second}. Future work will test the system on real-world scenarios.

\section*{Acknowledgment}
This work was supported by the Spanish projects PID2022-139615OB-I00, and DGA
FSE-T45 20R.

\balance
\bibliographystyle{IEEEtran}
\bibliography{biblio}

\end{document}